\documentclass[10pt,twocolumn,letterpaper]{article}

\usepackage{wacv}
\usepackage{times}
\usepackage{epsfig}
\usepackage{graphicx}
\usepackage{amsmath}
\usepackage{amssymb}
\usepackage[hang,flushmargin]{footmisc}
\usepackage{multirow}
\usepackage{color}
\wacvfinalcopy % *** Uncomment this line for the final submission

\ifwacvfinal
\def\assignedStartPage{1} % *** Enter the assigned starting page number (instead of 9876)
\fi

\ifwacvfinal
\usepackage[breaklinks=true,bookmarks=false]{hyperref}
\else
\usepackage[pagebackref=true,breaklinks=true,colorlinks,bookmarks=false]{hyperref}
\fi

\ifwacvfinal
\else
\fi

\begin{document}

%%%%%%%%% TITLE
\title{Distortion-aware Monocular Depth Estimation for Omnidirectional Images}
\author{Hong-Xiang Chen{$^\dagger$}, Kunhong Li{$^\dagger$}, Zhiheng Fu{$^\S$}, Mengyi Liu{$^\ddagger$}, Zonghao Chen{$^\ddagger$}, {Yulan Guo{$^\dagger$}\thanks{ indicates the corresponding author.}}\\
School of Electronics and Communication Engineering, Sun Yat-sen University\\
{\tt\small \{chenhx97, likh25\}@mail2.sysu.edu.cn, guoyulan@sysu.edu.cn}
\and
University of Western Australia{$^\S$}\\
{\tt\small{22907304@student.uwa.edu.au}}
\and
Alibaba Group{$^\ddagger$}\\
{\tt\small{\{suqing.lmy, czh190502\}@alibaba-inc.com}@alibaba-inc.com}}

\maketitle

%%%%%%%%% ABSTRACT
\begin{abstract}
    A main challenge for tasks on panorama lies in the distortion of objects among images. In this work, we propose a Distortion-Aware Monocular Omnidirectional (DAMO) dense depth estimation network to address this challenge on indoor panoramas with two steps. First, we introduce a distortion-aware module to extract calibrated semantic features from omnidirectional images. Specifically, we exploit deformable convolution to adjust its sampling grids to geometric variations of distorted objects on panoramas and then utilize a strip pooling module to sample against horizontal distortion introduced by inverse gnomonic projection. Second, we further introduce a plug-and-play spherical-aware weight matrix for our objective function to handle the uneven distribution of areas projected from a sphere. Experiments on the 360D dataset show that the proposed method can effectively extract semantic features from distorted panoramas and alleviate the supervision bias caused by distortion. It achieves state-of-the-art performance on the 360D dataset with high efficiency.
\end{abstract}

%%%%%%%%% BODY TEXT
\section{Introduction}

\begin{figure*}[t]
   \begin{center}
   \includegraphics[width=1.0\linewidth]{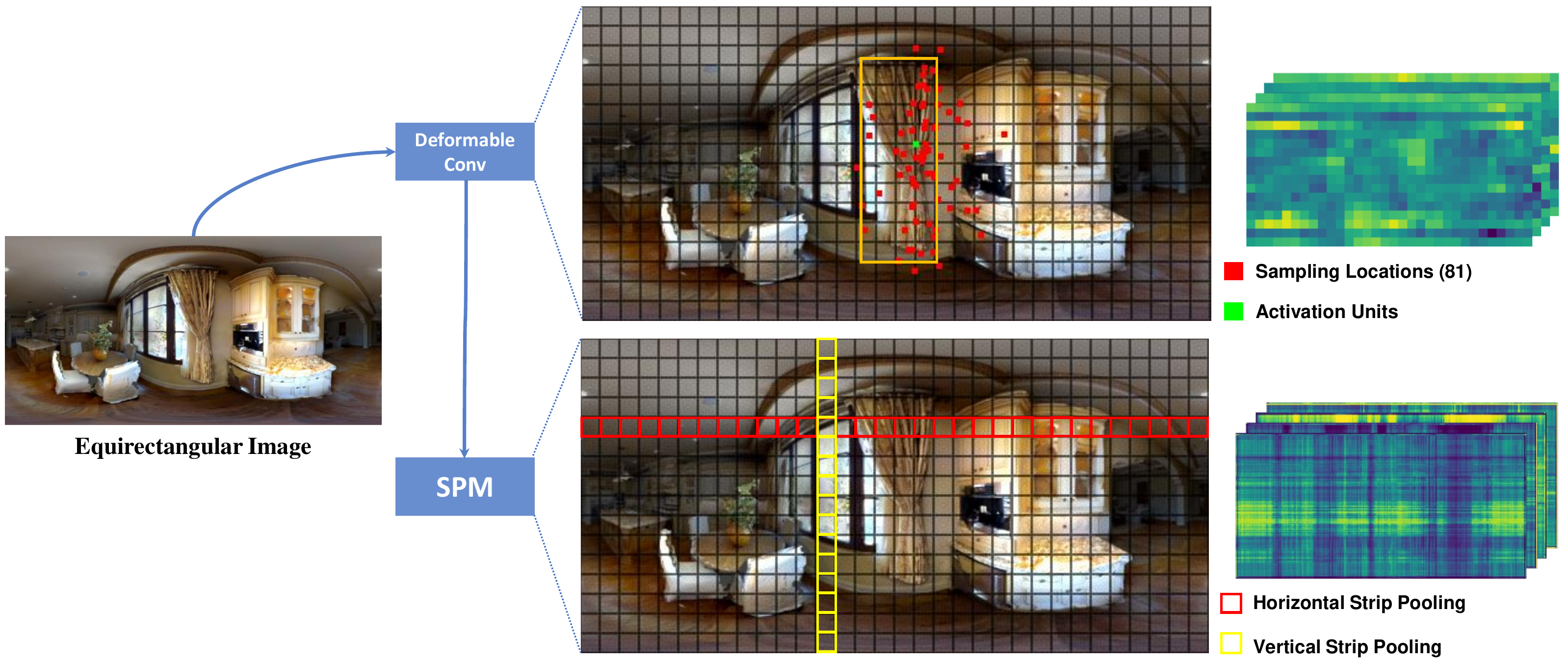} 
   \end{center} 
      \caption{The paradigm of our distortion-aware feature extraction block. To illustrate the learned distortion, we use the area around the curtain in the image as an example. $81$ locations (see red dots) are sampled by our DAMO network, it is clear that these sampling locations for an activation unit are mainly around the curtain (see green dot). Then, we utilize SPM to help deformable convolution focus mainly on informative regions and thus reduce the impact of distortion in panorama.}
   \label{fig:task}
\end{figure*}

%-------------------------------------------------------------------------

3D scene perception and understanding is a fundamental technique for many applications such as robotics and intelligent vehicles. Among various 3D vision
tasks, depth estimation is highly important since it forms the basis for many downstream tasks such as obstacle avoidance and object fetching. Due to the high costs of 3D sensors (e.g., LiDARs), inferring 3D information from 2D RGB images captured by cheap consumer-level cameras is significantly important.

Dense depth estimation is a challenging pixel-level task in 3D scene understanding. Different from stereo matching and Structure from Motion (SfM) methods, monocular depth estimation is an ill-posed problem for its information loss caused by the projection from a 3D space to a 2D image plane. That is, one pixel in a 2D image may correspond to multiple points in a 3D space. Thanks to the power of deep learning, progressive advances have been achieved using prior geometric constraints extracted either explicitly or implicitly from annotated data~\cite{laina2016deeper, eigen2015predicting, yin2019enforcing}. However, most of these methods focus on predicting depth from general perspective images.

Distortion is a major challenge for tasks working on panoramas, such as classification, saliency detection, semantic segmentation and depth estimation~\cite{eder2019convolutions, lee2019spherephd, zhang2018saliency}. Directly applying conventional CNNs on panoramas (e.g., represented in equirectangular formats) is hard to achieve promising performance. Since a panorama is usually produced by stitching several perspective images captured by a perspective camera located at the same place, equirectangular projection can be considered as a transformation from a non-Euclidean space to an Euclidean space and thus introducing distortion to panoramas. As a consequence, the projection of objects has irregular shapes and the distortion becomes extremely significant for pixels close to the poles or image plane. Therefore, standard convolution is unsuitable for panorama processing.

In this work, we propose a Distortion-Aware Monocular Omnidirectional (DAMO) network by combining strip pooling and deformable convolution to generate accurate depth maps from panoramas with distortion. We first present a distortion-aware feature extraction block to handle distortion introduced by equirectangular projection. Specifically, we utilize deformable convolution to learn offsets for the sampling grids, resulting in feature maps that are much denser than regular grids. We then exploit strip pooling to capture anisotropy context information from irregular regions (i.e., the distorted projection of objects) and preserve the integral distortion information for convolution sampling.  In addition, to mitigate supervision bias caused by uneven sampling in different areas, we also propose an easy-to-use spherical-aware weight matrix for the objective function. Experiments on the 360D dataset demonstrate that our DAMO network achieves the state-of-the-art performance with high efficiency.

Our contributions can be summarized as follows:

\begin{itemize}
\item We propose a DAMO network to handle distortion in panoramas using both deformable convolution and strip pooling module. Experiments on the 360D dataset show that DAMO is superior to the state-of-the-art.
\item We introduce a plug-and-play spherical-aware weight for our objective function to make the network focus on informative areas. This weight helps our network to achieve fast convergence and improved performance.
\end{itemize}

%-------------------------------------------------------------------------

\section{Related Work}

We will briefly describe several existing methods related to our work in this section.

%-------------------------------------------------------------------------

\subsection{Depth Estimation}

Depth estimation has been a hot topic for a long time. Early studies~\cite{scharstein2001a, rajagopalan2004depth, liang2019stereo} in this area focused on developing algorithms to generate point correspondences in stereo images. Different from these methods, Delage et al.~\cite{delage2006a} developed a Bayesian framework to perform 3D indoor reconstruction from one single perspective image based on a strong floor-wall assumption. Saxena et al.~\cite{saxena2005learning, saxena2009make3d} used Markov Random Fields (MRFs) to incorporate multiscale and global image features to predict depth from a single RGB image.

Eigen et al.~\cite{eigen2014depth} proposed the first deep learning based network. They used a multiscale convolutional architecture to predict results in a coarse-to-fine manner. Eigen et al.~\cite{eigen2015predicting} then adopted a multi-task training scheme to further improve the performance of their model. Laina et al.~\cite{laina2016deeper} proposed a regularization concerning loss and a uniform up-projection module for monocular depth estimation, which have been frequently used in subsequent methods. Fu et al.~\cite{fu2018deep} considered the depth estimation task as an ordinal regression problem by applying a spacing-increasing discretization strategy and a well-designed ordinal regression loss. Yin et al.~\cite{yin2019enforcing} improved the supervision capability of the objective function by randomly selecting a number of ternary points and producing a virtual plane for each ternary. With its geometric supervision, the virtual normal loss improves the convergence of the depth estimation model. However, all these methods focus on perspective images and may easily trapped into suboptimal results while being directly applied to panoramas.

\begin{figure*}[t]
    \begin{center}
    \includegraphics[width=1.0\linewidth]{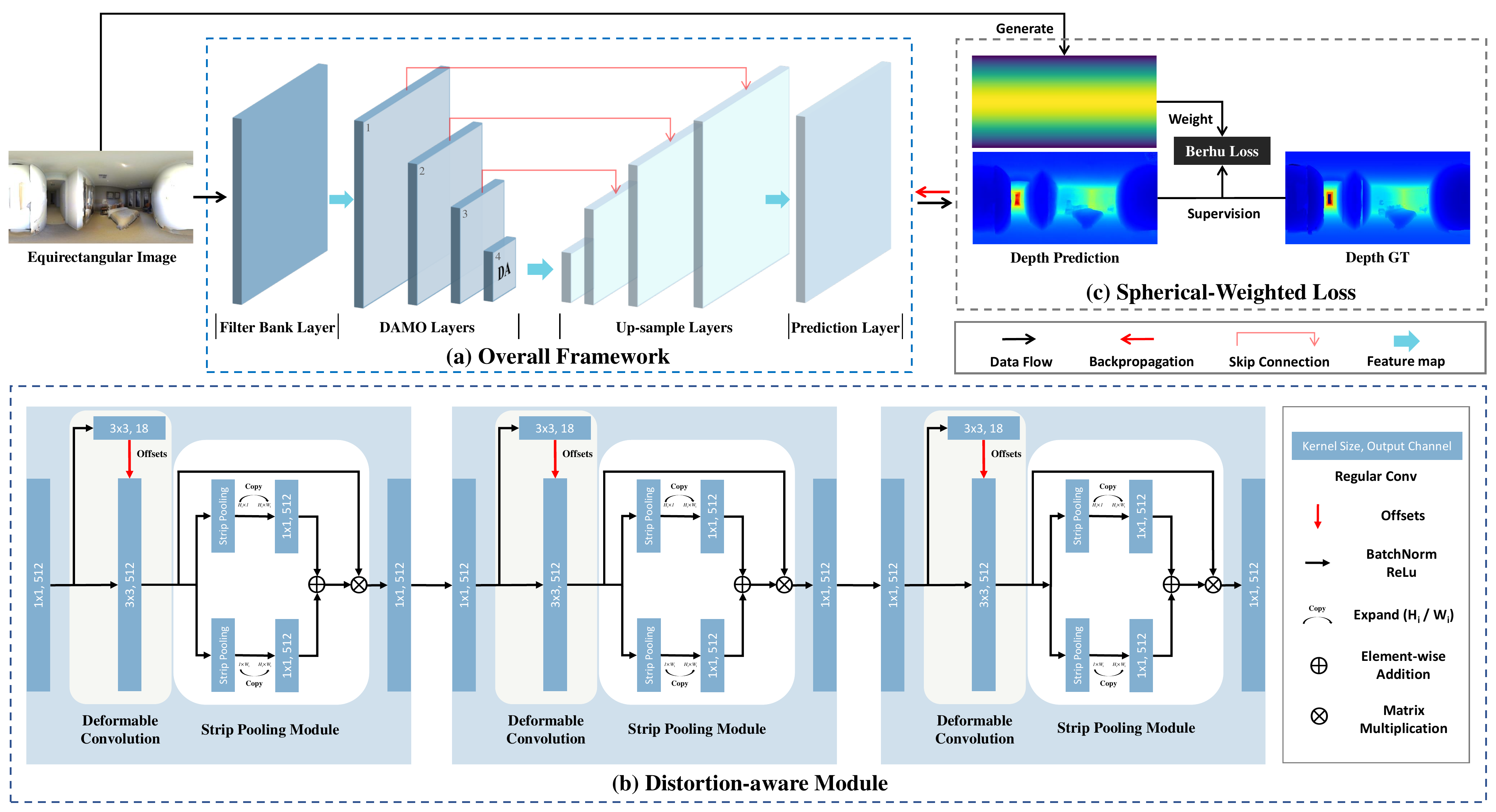}
    \end{center}
       \caption{An overview of our DAMO network. We build a filter-bank layer utilizing a group of parallel rectangular convolutions to extract horizontal distortion-aware features following~\cite{zioulis2018omnidepth,BiFuse20}. Then, these features are fed into an encoder-decoder based architecture~\cite{Ma2017SparseToDense} with a skip connection at each resolution. Besides, we also add SPM in the last building block for each of first three DAMO layers to help deformable convolution focus on contextual regions.}
       \label{fig:pipline}
 \end{figure*}

%-------------------------------------------------------------------------

\subsection{Representations on Panorama} \label{subsection:representations}

Equirectangular image is one of the most widely used representations for panoramas and distortion has been a major challenge for years. Su et al.~\cite{su2017flat2sphere:} used an adaptive kernel to handle the distortion near the pole. Following this idea, Zioulis et al.~\cite{zioulis2018omnidepth} designed a set of rectangular filter-banks to deal with the horizontal distortion introduced by equirectangular projection by increasing the receptive field of conventional convolution kernels on horizon. Although the representation capability of CNNs on panoramas has been improved by these methods, the gap between omnidirectional and perspective images still exists.

Cube map is another commonly used representation for panoramas. This representation faces the challenge of inconsistency between different faces. Cheng et al.~\cite{cubepadding2018} proposed cube padding to reduce the information loss along edges between faces. Wang et al.~\cite{BiFuse20} further extended~\cite{cubepadding2018} to spherical padding and propose a two-branch encoder-decoder based network to predict depth maps for panoramas. However, their model was hard to train due to its multiple training settings and high time cost. Inspired by rectangular convolution in~\cite{zioulis2018omnidepth}, we exploit strip pooling~\cite{hou2020strip} to preserve more context details for convolution.

%-------------------------------------------------------------------------

\subsection{Dynamic Mechanism} \label{dynamicmechanism} 

Existing deep learning based dynamic mechanisms can be divided into two categories: weight based methods~\cite{jia2016dynamic, hu2018squeeze, ma2020weightnet} and offset based methods~\cite{jeon2017active, dai2017deformable}.

Weight based methods focus on adaptively generating weights for either feature map selection or channel-wise selection. For instance, Jia et al.~\cite{jia2016dynamic} used a flexible filter generation network to produce a set of filter operators that dynamically conditioned on an individual input, resulting in improved performance on the video and stereo prediction tasks. Besides, attention is investigated to generate weights. Hu et al.~\cite{hu2018squeeze} proposed a light-weight gating mechanism to explicitly model channel-wise dependencies and further improve the model representation ability using global information.

Offset based methods aim at providing offsets for filters to aggregate more geometric information. Jeon et al.~\cite{jeon2017active} proposed an Active Convolution Unit (ACU) to learn its own shape adaptively during training. However, the shapes of filters have to be fixed after training. Moreover, Dai et al.~\cite{dai2017deformable} introduced deformable convolution to learn offsets for each spatial location dynamically, resulting in higher generalization capability than ACU.

%-------------------------------------------------------------------------
%-------------------------------------------------------------------------

\section{Proposed Method}

The overall pipeline of our DAMO network for monocular dense depth estimation on panorama is shown in Fig.~\ref{fig:pipline}. In this section, we will first introduce our Distortion-Aware (DA) module for calibrated semantic feature extraction, including  strip pooling  and deformable convolution. Then, we will introduce our objective function with a spherical-aware weight matrix.

%-------------------------------------------------------------------------

\subsection{Distortion-aware Module} \label{subsection:DA}

The DA module is shown in Fig.~\ref{fig:pipline}(b). We first utilize deformable convolution to extract learnable distorted information and calibrate semantic features. After that, the learned distortion knowledge is fed into the Strip Pooling Module (SPM)~\cite{hou2020strip}. The feature maps are further activated by multiple times and thus the distortion can be sufficiently learned by our DA module.

%-------------------------------------------------------------------------

\subsubsection{Deformable Convolution} \label{subsection:deformable}

To learn distortion knowledge and model the spatial transformation of convolution kernels on panorama, we adopt deformable convolution~\cite{dai2017deformable} in this work. Given a set of sampling locations on a regular grid $R$, the input feature map ${\bf{t}}$, the output feature map $F$ and the weights for kernel $\bf{w}$, a conv layer is applied in conventional convolution nerual networks. Here, we define the grid for $3 \times 3$ convolution kernel with a dilation of 1 as $R = \{(\pm{1},\pm{1}),(\pm{1},0),(0,\pm{1}),(0,0)\}$.

According to~\cite{dai2017deformable}, deformable convolution can be formulated as:

\begin{equation}
   \begin{aligned}
      {F({\boldsymbol{p_{0}}}) = \sum_{ {\boldsymbol{p_{n}}} \in R} {\bf{w}}( {\boldsymbol{p_{n}}}) \cdot {\bf{t}}({{\boldsymbol{p_{0}}} + {\boldsymbol{p_{n}}} + \Delta {\boldsymbol{p_{n}}}})}
   \end{aligned}
   \label{deformableconv}
\end{equation}

\noindent where ${\boldsymbol{p_{0}}}$ represents a location on output feature map $F$, ${\boldsymbol{p_{n}}}$ is one position on the regular sampling grid $R$ and $\Delta {\boldsymbol{p}_{n}}$ represents the offset corresponding to the position of ${\boldsymbol{p_{n}}}$.

The $i$-th input feature map (with a size of $C_{i} \times B_{i} \times H_{i} \times W_{i}$) is first fed into a convolutional layer to generate a group of 2D offsets for each sampling location on grid $R$. The offsets for both vertical and horizontal translation on $R$ have a size of $18 \times B_{i} \times H_{i} \times W_{i}$. Here, we use a $3 \times 3$ sampling grid $R$ as an example. Then, the sampling grid for deformable convolution is generated through Eq.~\ref{deformableconv} using the 2D offsets $\Delta \boldsymbol{p}_{n}$. Since most learnable offsets are fractional, we use bilinear interpolation to generate integer offsets ~\cite{dai2017deformable,zhu2019deformable}.

%-------------------------------------------------------------------------

\subsubsection{Strip Pooling} \label{subsection:strip}

To preserve more distortion information and help the network to focus on informative regions, we adopt strip pooling~\cite{hou2020strip} in our DA module. Different from standard spatial max pooling that adopts two-dimensional sampling grids, sampling along vertical and horizontal orientations are performed separately in strip pooling. That is, contexual information in feature maps are selected either in a row or a column according to its input (as shown in Fig.~\ref{fig:pipline}). Similar to the origin defination in~\cite{hou2020strip}, strip pooling is defined as:

\begin{equation}
    \begin{cases}
    y_{c, i}^{h} = {\max\limits_{0 \leq j < W}{x_{c,i,j}}} \\ 
    y_{c, j}^{v} = {\max\limits_{0 \leq i < H}{x_{c,i,j}}}
    \end{cases}
    \label{strip}
 \end{equation}
 
where $x_{c,i,j}$ is a location of the sampling grid on input (i.e., ${\bf{x}} \in \mathbb{R}^{C \times H \times W}$), $H$ and $W$ also represent the kernel size of strip pooling along horizontal and vertical orientations, respectively. ${y_{c,i}^{h} \in {\bf{y}^{h}}}$ and ${y_{c,j}^{v} \in {\bf{y}^{v}}}$ represent the $i$-th and $j$-th grids on the feature map of channel $c$ for horizontal and vertical strip pooling, respectively. 
 
Both vertical and horizontal strip pooling layers are used in SPM. Once these two expanded output feature maps (with the same size as their inputs) are added, an element-wise multiplication $G(\cdot,\cdot)$ is used to activate the contextual geometric areas of the input. SPM is sensitive to anisotropy context and produces denser distortion feature maps than the regular max pooling module. Given the combination ${\bf{y}}$ of ${\bf{y^{h}}}$ and ${\bf{y^{v}}}$, the output ${\bf{z}}$ of SPM  can be formulated as:

\begin{equation}
\begin{aligned}
   {{\bf{z}} = G({\bf{x}}, \sigma(f({\bf{y}})))}
\end{aligned}
\label{spm}
\end{equation}
where $\sigma$ is the sigmoid function, $f$ is the last $1 \times 1$ convolution layer, the fusion of horizontal and vertical strip pooling information ${y_{c,i,j} \in \bf{y}}$ is defined as ${y_{c,i,j} = y_{c,i}^{h} + y_{c,j}^{v}}$.

We apply our SPM to each conv$k$ layer of the DAMO layer (i.e., conv2, conv3, conv4 and conv5, as defined in ResNet~\cite{he2016deep}). Specifically, SPMs are used in the last building block of the first three DAMO layers and are stacked for all building blocks of the  DA module (see Fig.~\ref{fig:pipline}).

%-------------------------------------------------------------------------

\subsection{Spherical-aware Weighted Loss} \label{subsection:Spherical-Weighted}

Since an equitangular image represents a panorama in 2D space, distortion is extremely large in the areas around the pole of a spherical space. Specifically, for two areas with the same coverage in a spherical surface, the area near the pole is much larger than the area near the equator in an equitangular image due to uneven projection. Loss functions defined in perspective images~\cite{zioulis2018omnidepth, Eder_2020_CVPR, cheng2020depth} are hard to produce optimal results due to the overfitting (weighting) in sparse areas near the pole and the underfitting in dense areas near the equator.

%-------------------------------------------------------------------------

\subsubsection{Weight Matrix}

To achieve balanced supervision in different areas, we introduce a spherical-aware weighting strategy on objective function. Specifically, the Cartesian coordinates of a pixel ${p_{E} = D(x,y)}$ in the equirectangular image can be converted to spherical coordinates ${p_{S} = \Pi(\theta, \phi)}$ in a spherical surface, where longtitude ${\theta \in [0 , 2\pi]}$ and latitude ${\phi \in [0, \pi]}$. That is, $\phi_{(x,y)} = \frac{\pi x}{\rm{H}}, \theta_{(x,y)} = \frac{2 \pi y}{\rm{W}}$.

Considering a sphere $\Pi(\theta, \phi)$ with a unit radius, we generate a weight matrix for objective functions according to the sphere angle of a pixel along the latitude. Take the north hemisphere as an example, the weight is defined as the ratio of the area from the north pole to the current latitude to the total area of the sphere surface. The weights in the south hemisphere can be calculated following a similar way.

\begin{equation}
    \begin{aligned}
       {\mathrm{W}_{(x,y)} =  {\int_{0}^{\phi_{(x,y)}}{{\rm{sin}}{\phi} d_{\phi}}}}
    \end{aligned}
    \label{weight}
\end{equation}
 
Here, $\mathrm{W}$ is the weight matrix for the objective function in each location. $\phi_{(x,y)}$ denotes the angle of the position $(x,y)$ along the vertical axis.

\begin{table*}
    \centering
    \caption{Comparison to the state-of-the-art $360^{\circ}$ monocular depth estimation methods on the 360D Dataset. The method with $^{\divideontimes}$ represents a model provided by the author (with its Caffe based weights being converted to Pytorch based weights), and the method with $^{\star}$ denotes our reproduction. Note that, the results and metrics reported in~\cite{BiFuse20} are different from~\cite{zioulis2018omnidepth}, to follow the baseline method of the 360D Dataset, we convert its \textit{RMSE(log)} results from base-10 logarithm to natural logarithm. Besides, the results in~\cite{zioulis2018omnidepth} are updated at the authors' github repository$^{1}$.} 
    \label{table:360D}
    \begin{tabular}{|c|c|c|c|c|c|c|} 
    \hline  
    {\multirow{2}{*}{\textbf{Method}}}&\textbf{RMSE}&\textbf{Abs\_REL}&\textbf{RMSE(log)}&\textbf{$\delta_{1}$}&\textbf{$\delta_{2}$}&\textbf{$\delta_{3}$}\\
    \cline{2-7}
    ~&\multicolumn{3}{c|}{Lower the better}&\multicolumn{3}{c|}{Higher the better}\\
    \hline
   OmniDepth-UResNet~\cite{zioulis2018omnidepth}   & 0.3084 & 0.0946 & 0.1315 & 0.9133 & 0.9861 & 0.9962 \\
   OmniDepth-RectNet~\cite{zioulis2018omnidepth}   & 0.2432 & 0.0687 & 0.0999 & 0.9583 & 0.9936 & 0.9980 \\
    \hline
    BiFuse-Equi~\cite{BiFuse20} & 0.2667 & - & 0.1006 & 0.9667 & 0.9920 & 0.9966  \\
    BiFuse-Cube~\cite{BiFuse20} & 0.2739 & - & 0.1029 & 0.9688 & 0.9908 & 0.9956  \\
    BiFuse-Fusion~\cite{BiFuse20} & 0.2440 & - & \textit{0.0985} & \textit{0.9699} & 0.9927 & 0.9969  \\
    \hline
    OmniDepth-RectNet${}^{\divideontimes}$ & \textit{0.2297} & 0.0641 & 0.0993 & 0.9663 & \textit{0.9951} & \textit{0.9984} \\
    \hline
    BiFuse-Equi${}^{\star}$ & 0.2415 & \textit{0.0573} & 0.1000 & 0.9681 & 0.9928 & 0.9972  \\
    \hline
    DAMO & \textbf{0.1769} & \textbf{0.0406} & \textbf{0.0733} & \textbf{0.9865}& \textbf{0.9966} & \textbf{0.9987} \\ 
   \hline
    \end{tabular}
\end{table*}

%-------------------------------------------------------------------------

\subsubsection{Loss Function}

Following~\cite{yin2019enforcing, fang2020towards}, we adopt the reverse Huber loss (also called the Berhu loss)~\cite{laina2016deeper} in this work.

\begin{equation}
    \mathcal{L}_{d}=
    \begin{cases}
    \left|d{_{pre}}-d{_{gt}}\right|,  & \text{if $\left|d{_{pre}}-d{_{gt}}\right| \leq \tau$} \\
    \frac {(d{_{pre}}-d{_{gt}})^2+\tau^2}{2\tau}, & \text{if $\left|d{_{pre}}-d{_{gt}}\right| > \tau$}
    \end{cases}
    \label{berhu}
 \end{equation}

\noindent where $d_{pre}$ and $d_{gt}$ are the predicted and groundtruth depth values, respectively. The Berhu loss can achieve a good balance between L1 and L2 norms. Specifically, pixels with high gradient residuals will be assigned with large weights using the L2 term. Meanwhile, the L1 term pays more attention to regions with small gradients. In our experiments, we set the threshold $\tau$ as $20\%$ of the maximum error between prediction and groundtruth. 

Finally, our loss function is defined as:

\begin{equation}
    {\mathcal{L} = \mathcal{L}_{d} \otimes \mathrm{W} }
    \label{weight-berhu}
\end{equation}

Compared to the original Berhu loss $\mathcal{L}_{d}$, our weighted Berhu loss $\mathcal{L}$ can help the network to focus more on informative regions (i.e, regions near the equator) and mitigate the effects introduced by severe distortion (which is significant in areas near the pole) in panoramas.

%-------------------------------------------------------------------------
\begin{figure*}[t]
    \begin{center}
    \includegraphics[width=1.0\linewidth]{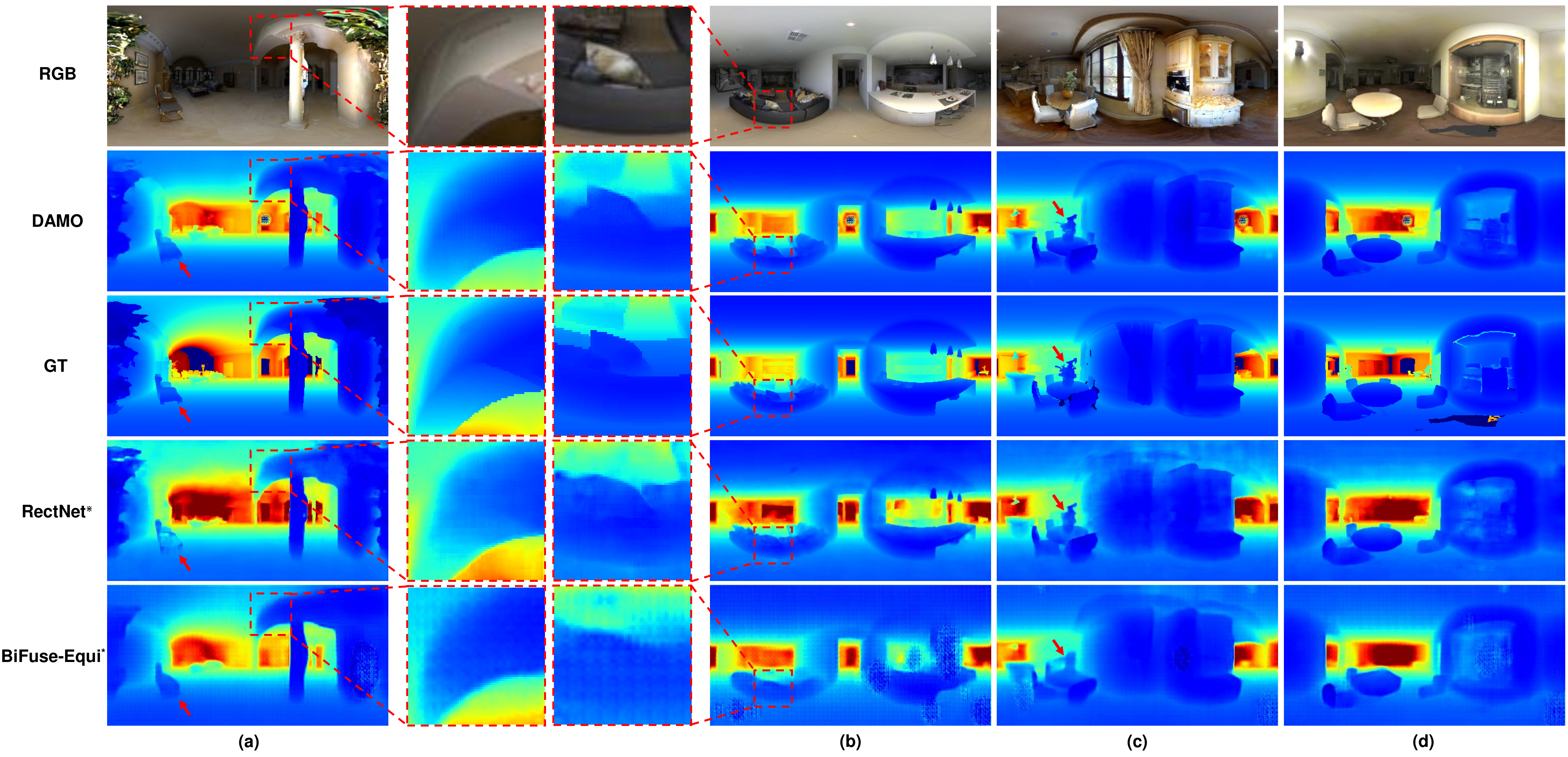}
    \end{center}
       \caption{Qualitative Comparison on the 360D Dataset. Here, we colorize these depth maps to better distinguish the effectiveness of different methods. Points with dark color (blue) are closer than those with light color (red). The first row shows equirectangular RGB images, while invalid areas are masked with black in groundtruth (in the third row). It is obvious that our DAMO generates more accurate depth maps than existing methods, especially for those distorted areas.}
    \label{fig:com1}
\end{figure*}
%-------------------------------------------------------------------------

%-------------------------------------------------------------------------
%-------------------------------------------------------------------------

\section{Experiments} \label{section:Experiments}

We conduct extensive experiments on a widely used omnidirectional dataset to evaluate the performance of our DAMO network. We will first describe the dataset and the evaluation toolbox, and then present the implementation details of our experiments. We further compare our method to the state-of-the-art.

%-------------------------------------------------------------------------

\subsection{Experimental Settings}

\subsubsection{Dataset}

We adopt a large-scale indoor omnidirectional RGBD dataset (i.e., the 360D Dataset~\cite{zioulis2018omnidepth}) to conduct experiments. This dataset contains two real-world datasets (i.e., Stanford2D3D~\cite{Armeni2017Joint} and Matterport3D~\cite{Matterport3D}), and two synthetic datasets (i.e., SunCG~\cite{song2017semantic} and SceneNet~\cite{handa2016scenenet}). Following the original split in ~\cite{zioulis2018omnidepth}, the training and test sets are listed as follows:

\begin{itemize}
    \item  Training set: a cross-domain set of both real-world and sythetic images from the Stanford2D3D, Matterport3D and SunCG datasets were first obtained. Then, scenes with very close or far regions were removed, resulting in a training dataset with 34,679 RGBD image pairs.
    \item Test set: 1,289 omnidirectional image pairs were collected from the Stanford2D3D, Matterport3D and SunCG dataset for test. The remaining image pairs from the SceneNet dataset were used for validation.
\end{itemize}

%-------------------------------------------------------------------------

\subsubsection{Implementation Details}\label{subsubsection:implement}

Our network was implemented in Pytorch with a single Nvidia RTX Titan GPU. Each RGB image has a resolution of $512 \times 256$ while invalid depth values are removed by a mask. The batch size was set to 8 and the initial learning rate was set to $1 \times 10^{-4}$. We used the Adam optimizer~\cite{kingma2014adam} with its default settings and a poly learning rate policy~\cite{fu2019dual} for training. All models were trained for 20 epochs on the 360D dataset for fair comparision.

%-------------------------------------------------------------------------

\subsubsection{Evaluation Metrics} \label{subsubsection:Metrics}

We adopted the same metrics as previouse works~\cite{zioulis2018omnidepth, cheng2020depth} for fair comparison, including Absolute average Relative Error (Abs\_REL), Root Mean Squared Error (RMSE), Root Mean Squared Error in logarithmic space (RMSElog) and accuracy with a threshold $\delta_{t}$, where ${t \in \{1.25, 1.25^{2}, 1.25^{3} \} }$. Note that, we used the same evaluation strategy as ~\cite{zioulis2018omnidepth}. That is, depth maps were estimated by dividing a median scalar $\bar{s}$ to achieve direct comparison among multiple datasets with different range scales, where $\bar{s} = {\rm{median}}{(D_{GT})} / {\rm{median}}{(D_{Pred})}$.

%-------------------------------------------------------------------------

\subsection{Comparison to the State-of-the-art} \label{subsubsection:sota}

We compare our DAMO to two existing methods. Note that, only the Caffe model is provided by~\cite{zioulis2018omnidepth} while the source codes are provided by~\cite{BiFuse20}. We used their released model or source codes for comparison in this work.
%-------------------------------------------------------------------------

\subsubsection{Quantitative Comparison}

As shown in Table~\ref{table:360D}, our method outperforms the baseline method~\cite{zioulis2018omnidepth} by a large margin. Specifically, the {Abs{\_}Rel} and {RMSE} values of DAMO are much better than 
OmniDepth-RectNet$^{\divideontimes}$ by $\textbf{36.5\%}$ and $\textbf{22.9\%}$, respectively. Although the same backbone (i.e., ResNet-50~\cite{he2016deep}) is used in BiFuse-Equi~\cite{BiFuse20}, BiFuse-Fusion~\cite{BiFuse20}, and our DAMO model, our model achieves significant performance improvement over the BiFuse-Equi and BiFuse-Fusion methods in almost all metrics. Note that, the BiFuse-Fusion method adopts an additional cube map representation branch and thus has more parameters to tune. Besides, the cube map representation has obvious discontinuity between neighboring faces. Although different padding schemes have been proposed to mitigate the edge effect~\cite{cubepadding2018, BiFuse20}, the computational efficiency of this representation is still low. In contrast, our DAMO network has only one equirectangular branch to predict depth on panorama and outperforms BiFuse-Fusion by nearly $\textbf{26.7\%}$ in RMSE.

\footnotetext[1]{https//github.com/VCL3D/360Vision/tree/master/SingleImageDepthMetrics.}

The superiority of our DAMO network can be attributed to two reasons. \textbf{First}, DA module can adjust irregular sampling grids among distorted projection of objects in panoramas automatically and extracts rich geometric information in challenging areas (e.g., pole areas). \textbf{Second}, our weighted Berhu loss can help our model focusing on informative areas (especially for these areas near the equator) and alleviate the supervision bias caused by distortion. Consequently, the representation capability of our network on panoramas is improved.

%-------------------------------------------------------------------------

\begin{figure*}[t]
    \begin{center}
    \includegraphics[width=1.0\linewidth]{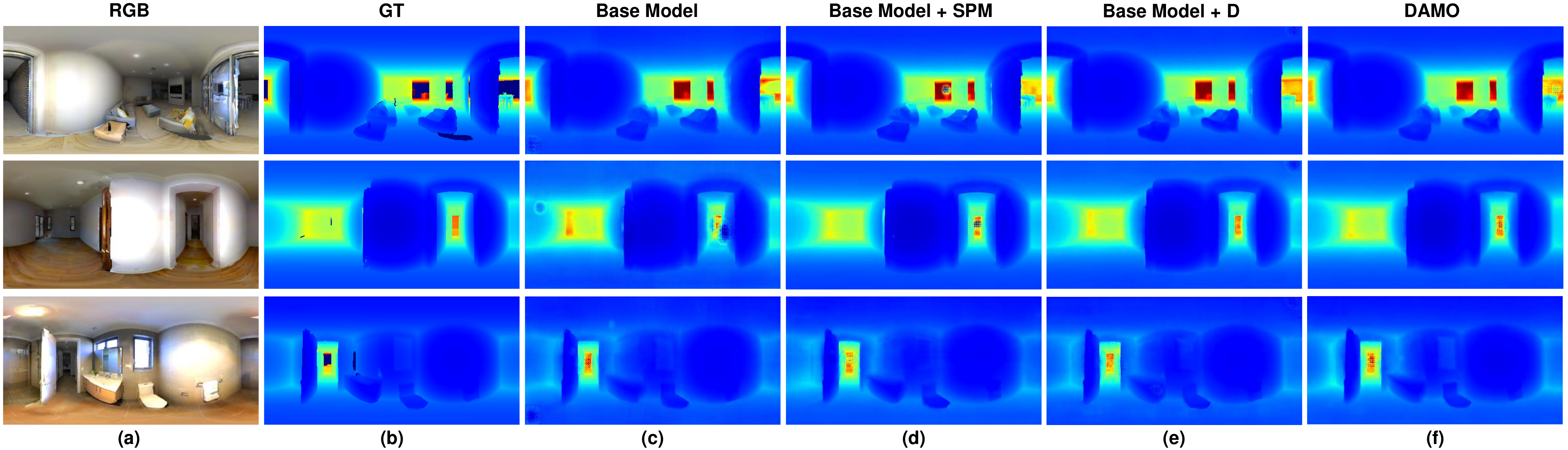}
    \end{center}
       \caption{Ablation study with strip pooling and deformable convolution on the 360D Dataset. Models with strip pooling can predict clearer depth regions than the base model, while models with deformable convolution generate much sharper boundaries than those with regular convolution. Integrating both strip pooling and deformable convolution in our DA module can further improve the performance.}
    \label{fig:com2}
\end{figure*}
%-------------------------------------------------------------------------

\subsubsection{Qualitative Comparison} 
We present several predicted depth maps from the 360D dataset in Fig.~\ref{fig:com1}. It can be observed from the zoom-in regions that DAMO can predict more accurate and clearer depth maps than RectNet~\cite{zioulis2018omnidepth} and BiFuse-Equi~\cite{BiFuse20} on panoramas. It is also shown that fine details such as chair and vase (as denoted by a small arrow in Fig.~\ref{fig:com1}) can be successfully estimated. This further demonstrates the representation capability of our DAMO network on various types of objects in equirectangular images.

%-------------------------------------------------------------------------
%-------------------------------------------------------------------------

\setlength{\tabcolsep}{1mm}{
\begin{table}
   \centering
   \caption{Ablation study with strip pooling and deformable convolution. The second and third parts show the performance achieved by networks with SPM and deformable convolution (where `D' represents deformable convolution), respectively. $\mathcal{L}$ indicates the spherial-weighted loss. All models are trained in the same settings and the best results in each part are shown in boldface.}
   \label{table:ablations}
   \begin{tabular}{|c|c|c|c|c|} 
   \hline  
   {\multirow{2}{*}{\textbf{Method}}}&{\multirow{2}{*}{\textbf{Param.}}}&\textbf{RMSE}&\textbf{Abs\_REL}&\textbf{RMSE(log)}\\ 
   \cline{3-5}
   ~&~&\multicolumn{3}{c|}{Lower the better}\\ 
   \hline
   Base Model & {\multirow{2}{*}{\textbf{61.28M}}} &0.2129 & 0.0598 & 0.0959 \\ 
   + $\mathcal{L}$ &~& \textbf{0.2068} & \textbf{0.0562} & \textbf{0.0904} \\ 
   \hline 
   + SPM & {\multirow{2}{*}{+ \textbf{5.83M}}} & 0.1879 & 0.0433 & 0.0769 \\ 
   + $\mathcal{L}$ &~& \textbf{0.1803} & \textbf{0.0419} & \textbf{0.0749} \\ 
    \hline 
   + D (conv5) & {\multirow{2}{*}{+ \textbf{0.24M}}} & 0.1906 & 0.0471 & \textbf{0.0789} \\ 
   + $\mathcal{L}$ &~& \textbf{0.1882} & \textbf{0.0455} & 0.0794 \\   
    \hline
   \end{tabular}
\end{table}
}
  
\begin{figure*}[t]
   \begin{center}
   \includegraphics[width=1.0\linewidth]{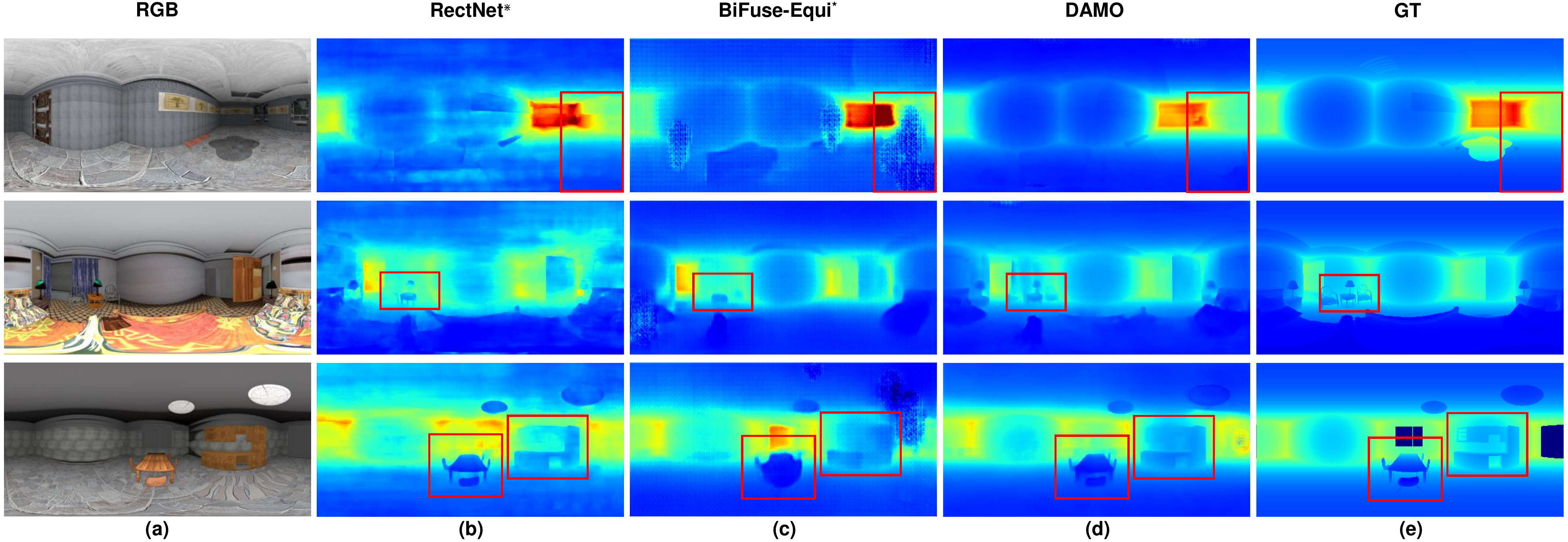}
   \end{center}
      \caption{Generalization analysis for the proposed methods. We evaluate existing methods in SceneNet to test the generalization ability of the model.}
   \label{fig:scannet}
\end{figure*}
%-------------------------------------------------------------------------
%-------------------------------------------------------------------------

\subsection{Ablation Study} \label{subsubsection:ablation}

In this section, we conduct experiments to illustrate the effectiveness of each component of our DAMO network.

%-------------------------------------------------------------------------

\subsubsection{Effectiveness of SPM}

We will first analyze the benefit of exploiting SPM in our DA module. As illustrated in Table~\ref{table:ablations}, it can be observed that the improvement of applying SPM is significant. While the number of parameters of the model is increased by {$\textbf{5.83}$M} (nearly $9.5\%$ as compared to the base model), the performance is improved by about $\textbf{27.5\%}$ and $\textbf{11.7\%}$ in {Abs\_REL} and {RMSE}, respectively. We argue that the trade-off between the complexity of network and advancement of its performance is well balanced. As shown in Fig.\ref{fig:com2}, the base model produces many artifacts, especially on the south and north poles. In contrast, the base model with SPM can extract rich contextual information and predicts fine depth map on panoramas. 

%-------------------------------------------------------------------------

\subsubsection{Deformable vs. Regular Convolution} 

We compare deformable convolution to its regular counterpart in this section. As shown in Table~\ref{table:ablations}, the performance of our baseline is improved significantly with deformable convolution. Specifically, {Abs{\_}REL} is improved from $0.0598$ to $0.0471$ (nearly by $\textbf{21.2\%}$) and RMSE is improved from $0.2129$ to $0.1906$ (nearly by $\textbf{10.4\%}$) by integrating deformable convolution into our DAMO network.

As shown in Figs.~\ref{fig:com2}(c)(e), we can observe that the model with deformable convolution can generate much cleaner depth maps on walls, ceilings and floors. Note that, distortion exists mainly in these areas since ceilings and floors are always located in the north and south pole in images of the 360D dataset. Besides, long walls usually exist from left to right in indoor scenes. It is clear that deformable convolution can learn reasonable offsets to model the transformation of sampling grids and mitigate the harmful distortion effects for CNNs introduced in panoramas. Moreover, deformable convolution can produce sharper object boundaries and more accurate depth results in some difficult regions (e.g., see Fig.~\ref{fig:com2}(e)). 

%-------------------------------------------------------------------------

\subsubsection{Deformable Convolution with SPM}

Here, we investigate how these two components of our DA module work together in synergy. As shown in Tables~\ref{table:360D} and \ref{table:ablations}, although deformable convolution or SPM improves our base model for more than $\textbf{10.0\%}$ in {Abs\_REL} and {RMSE}, the improvement of their combination is still significant. Specifically, our DAMO network outperforms the base model by nearly $\textbf{31.9\%}$ in {Abs\_REL} and $\textbf{16.9\%}$ in {RMSE}. We believe the main reason for our improvement on panorama is that the DA module learns the distortion in this domain. The synergism of SPM and deformable convolution is obvious. That is, SPM activates informative regions and helps deformable convolution to focus on challenging areas by learning the transformation of sampling grids among panoramas.

%-------------------------------------------------------------------------

\subsubsection{Spherical-aware Weight vs. Regular Weight}  \label{subsubsection:weight}

We conduct ablation experiments to demonstrate the effectiveness of our spherical-aware weight for objective functions. Table~\ref{table:ablations} show the results of each component with/without the proposed weight. The largest improvement is achieved on our base model, its {Abs{\_}REL} is improved from 0.0598 to 0.0562 (by nearly $\textbf{6.0\%}$) and RMSE is improved from 0.2129 to 0.2068 (by nearly $\textbf{2.8\%}$). When SPM or deformable convolution is used in the model, the spherical-aware weight can still introduce further performance improvement.
%-------------------------------------------------------------------------

\subsubsection{Generalization Analysis}

To further demonstrate the generalization capability of our DAMO network, we tested our method without finetune on the validation split of the 360D dataset (i.e., SceneNet~\cite{handa2016scenenet}). It is clear that DAMO outperforms other methods. As shown in Fig.~\ref{fig:scannet}, DAMO predicts more fine details and its results are closer to groundtruth than other methods. Compared to the test set of 360D, the overall performance drop of our DAMO network on the validation split is lower than other methods. For instance, our RMSE is increased by $0.1088$. In contrast, the RMSEs of BiFuse-Equi and OmniDepth-RectNet  are increased by $0.1437$ and $0.1471$, respectively. That is because, RectNet and BiFuse-Equi do not consider the distortion of panorama and are therefore overfitted on the training set of 360D. More quantitative details are reported in our supplementary material.

The superiority of DAMO can be attributed to our DA module, which consists of SPM and deformable convolution. Specifically, reasonable offsets are learned with deformable convolution and its sampling grids can target distorted projection of objects at different locations of a panorama. Therefore, our DAMO obtains transformation capability of sampling grids against distortion. Besides, the element-wise multiplication in SPM generates the connection among different channels in each input. Consequently, the transferability of DAMO is also improved. The DA module improves the overall representation capability of our network, especially along occlusion boundaries of objects (see the frame in Fig.~\ref{fig:scannet}).

%-------------------------------------------------------------------------
%-------------------------------------------------------------------------

\section{Conclusion}

We presented a method for omnidirectional dense depth estimation. We introduce deformable convolution to handle distortion of panorama and use strip pooling to improve the generalization ability of our DAMO network. To alleviate the supervision bias caused by distortion, we further introduce a spherical-weighted objective function to aggregate abundant information near the equator of the sphere. Experiments show that the proposed method outperforms the state-of-the-art monocular omnidirectional depth estimation methods. 

%-------------------------------------------------------------------------
%-------------------------------------------------------------------------

{\small
\bibliography{damo}
\bibliographystyle{ieee_fullname}
}

\end{document}